\documentclass{article}
\usepackage{ifthen}
\newboolean{showcomments}
\setboolean{showcomments}{false}
\usepackage{float}

\newcommand{\grace}[1]{\ifthenelse{\boolean{showcomments}}{\textcolor{magenta}{grace: #1}}{}}
\newcommand{\raj}[1]{\ifthenelse{\boolean{showcomments}}{\textcolor{orange}{raj: #1}}{}}
\newcommand{\jenna}[1]{\ifthenelse{\boolean{showcomments}}{\textcolor{green}{jenna: #1}}{}}
\newcommand{\sv}[1]{\ifthenelse{\boolean{showcomments}}{\textcolor{blue}{sv: #1}}{}}
\usepackage{listings}
\usepackage[T1]{fontenc}
\usepackage{tikz}
\usetikzlibrary{fit,tikzmark}
 
\usepackage{enumitem}

\usepackage[round, sort, numbers]{natbib}

\usepackage{amsmath}
\usepackage{graphicx}
\usepackage{subfigure}
\usepackage{subcaption}

\usepackage{pifont}
\usepackage{amssymb}
\newcommand{\lt}{<}
\newcommand{\gt}{>}

\newcommand{\cmark}{\textcolor{green!80!black}{\ding{51}}}
\newcommand{\xmark}{\textcolor{red}{\ding{55}}}

    \usepackage[final]{neurips_2024}

\usepackage[utf8]{inputenc} 
\usepackage[T1]{fontenc}    
\usepackage{hyperref}       
\usepackage{url}            
\usepackage{booktabs}       
\usepackage{amsfonts}       
\usepackage{nicefrac}       
\usepackage{microtype}      
\usepackage{xcolor}         

\title{Understanding Graphical Perception in Data Visualization through Zero-shot Prompting of Vision-Language Models}

\newcommand{\gtlogo}{\raisebox{3.4pt}{\includegraphics[scale=0.036]{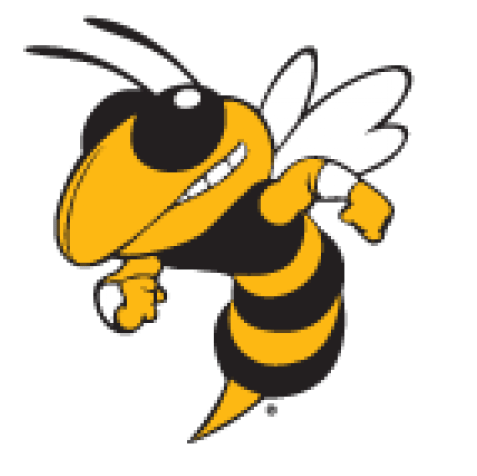}}}
\newcommand{\nyulogo}{\raisebox{3.4pt}{\includegraphics[scale=0.06]{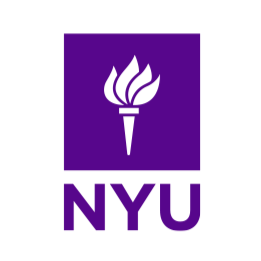}}}
\newcommand{\hrvlogo}{\raisebox{3.4pt}{\includegraphics[scale=0.05]{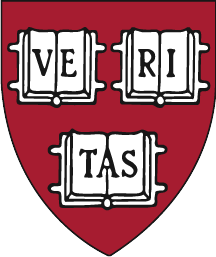}}}

\makeatletter
\def\thanks#1{\protected@xdef\@thanks{\@thanks
        \protect\footnotetext{#1}}}
\makeatother
\author{ 
        Grace Guo  \hrvlogo \footnotemark[1] \thanks{* Equal contribution.} ,
        Jenna Jiayi Kang \nyulogo \footnotemark[1] ,
        Raj Sanjay Shah \gtlogo \footnotemark[1] ,\\
       Hanspeter Pfister \hrvlogo,
       Sashank Varma \gtlogo\\
       Harvard University \hrvlogo, 
        New York University \nyulogo, 
        Georgia Institute of Technology \gtlogo 
        \thanks{Email: gguo31@g.harvard.edu,  jennakang@nyu.edu, rajsanjayshah@gatech.edu, pfister@seas.harvard.edu, varma@gatech.edu}
        }

\begin{document}

\maketitle

\begin{abstract}
Vision Language Models (VLMs) have been successful at many chart comprehension tasks that require attending to both the images of charts and their accompanying textual descriptions.
However, it is not well established how VLM performance profiles map to human-like behaviors.
If VLMs can be shown to have human-like chart comprehension abilities, they can then be applied to a broader range of tasks, such as designing and evaluating visualizations for human readers.
This paper lays the foundations for such applications by evaluating the accuracy of zero-shot prompting of VLMs on graphical perception tasks with established human performance profiles.
Our findings reveal that VLMs perform similarly to humans under specific task and style combinations, suggesting that they have the potential to be used for modeling human performance.
Additionally, variations to the input stimuli show that VLM accuracy is sensitive to stylistic changes such as fill color and chart contiguity, even when the underlying data and data mappings are the same.
\end{abstract}

\section{Introduction and Related Work}



Vision Language Models (VLMs) are capable of synthesizing information in both the vision and language input modalities, leading to their application in healthcare diagnostics \cite{soenksen2022integrated}, autonomous vehicles \cite{nie2020multimodality}, interactive robotic applications \cite{zhang2024mm}, and other domains.
In our domain of interest, \emph{data visualization}, VLMs have also been used for a range of tasks that require attending to both the images of charts and graphs and their accompanying textual descriptions \cite{han2023chartllama, meng2024chartassisstant, masry2024chartgemma, masry2024chartinstruct, xu2023chartbench}, from simple tasks such as data extraction \cite{meng2024chartassisstant} and question answering \cite{xia2024chartx, meng2024chartassisstant, masry2024chartgemma, masry2024chartinstruct,masry2022chartqa, kahou2017figureqa, kafle2018dvqa, methani2020plotqa, chaudhry2020leaf, singh2020stl, kantharaj2022opencqa, wan2024datavist5, masry2023unichart} to more complex tasks such as chart generation and refinement \cite{han2023chartllama}.
Recent research has evaluated whether VLMs show human-like visualization comprehension abilities using visualization literacy tests \cite{bendeck2024empirical}. Such tests consist of questions that measure the ability of humans to comprehend and extract information from visualizations. 
Studies with GPT-4 show that it can reason about visualizations, identify trends, and suggest best design practices. Yet, the model struggles with simple tasks like value retrieval and color distinctions in charts.
If VLMs show human-like visualization comprehension abilities, they can be used to design and evaluate visualizations, e.g., identifying potential sources of cognitive processing (over)load. 
However, doing so requires establishing that VLM performance profiles map to human-like behaviors. \emph{Here, we lay the foundations for such applications by evaluating the accuracy of VLMs when performing graphical perception tasks.}

\begin{figure}
    \centering
    \includegraphics[width=\linewidth]{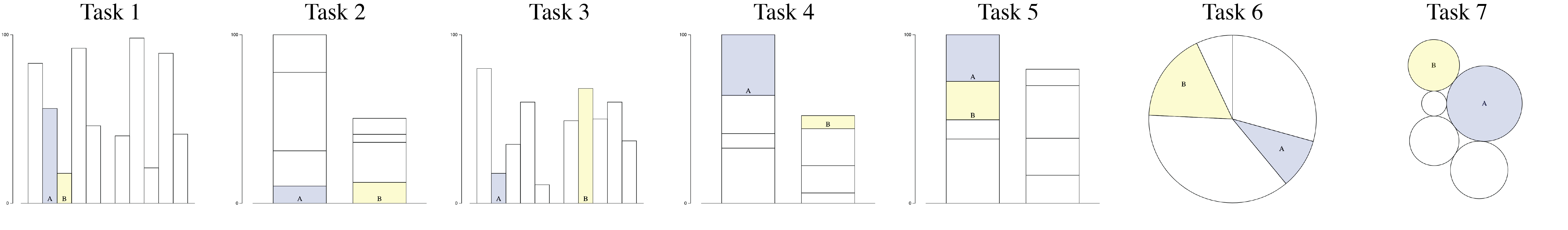}
    \caption{Examples of the seven tasks in our study, adapted from \cite{heer2010crowdsourcing}.
    For each visualization, the VLM was prompted to compare the two segments in blue and yellow (also labeled A and B, respectively).
    }
    \label{fig:tasks}
\end{figure}

Graphical perception tasks require elementary perceptual operations such as retrieving numerical values from positional encodings, lengths, and angles. They were introduced by Cleveland and McGill in 1984 \cite{cleveland1984graphical}, in a series of experiments where participants were asked to extract two numerical quantities from a chart and to judge the proportion of the smaller quantity against the larger (see Tasks 1-6 of Fig. \ref{fig:tasks}). \citet{heer2010crowdsourcing} later replicated this study with a larger pool of participants recruited from the crowd-sourcing platform MTurk. They also extended the stimuli to include other types of judgment tasks, such as area judgments (Fig. \ref{fig:tasks}, Task 7). \emph{These studies revealed potential sources of cognitive processing load in the complexity of common visualizations.} 
Inspired by these studies, other prior work have used the same stimuli to investigate relational reasoning in CNNs compared to human performance \cite{cui2024generalization}.  We extend these experiments to evaluate the human-like performance of popular VLMs in a zero-shot, out-of-the-box manner.



In this work, we evaluate whether VLMs can simulate human graphical perception performance when performing the same seven tasks from these seminal studies \cite{cleveland1984graphical, heer2010crowdsourcing}. To do so, we first recreated the original stimuli, implementing 45 trials for each of the seven chart types shown in Figure \ref{fig:tasks}. Each trial included a visualization with two segments highlighted. We then \emph{zero-shot prompted} the GPT-4o-mini model \cite{openai2023gpt4} to 1) indicate which segment is smaller, and 2) estimated percentage of the smaller segment is the larger, in a procedure similar to the one used by \citet{heer2010crowdsourcing}.
Overall, our contributions are:
\begin{itemize}[leftmargin=*]
    \setlength\itemsep{0em}
    \setlength\parskip{0em}
    \setlength\parsep{-2em}
 \item \textbf{Behavioral evaluation of VLMs on graphical perception tasks:} We assess whether GPT-4o-mini can simulate human-like behaviors by comparing the accuracy and confidence of the VLM in interpreting visualizations against human performance profiles \cite{cleveland1984graphical, heer2010crowdsourcing}.
 \item \textbf{Model performance across prompts:} We use four prompt variations to test the suitability of VLMs for modeling human graphical perception -- with and without references to the target segment colors, and with and without generation of explanations/reasonings in the output template \cite{wei2022chain}.
 \item \textbf{Model performance across stimuli:} We introduce variations in the stimuli as shown in Figure \ref{fig:study_variations}a to test how incidental factors influence the model's performance in interpreting visual data.
 \item \textbf{Model performance on new tasks:} We implement novel task variants, shown in Figure \ref{fig:study_variations}b, and evaluate whether VLMs show a performance decrement when the critical elements are contiguous.
\end{itemize}



\section{Method}

Our work adapts the stimuli and tasks from two prior human studies to evaluate the behavioral alignment of the graphical perception abilities of VLMs. To ensure the comparability of results across studies, we recreate the stimuli and prompt the VLM with the same probes in a zero-shot manner. Information for stimulus generation was taken from both \cite{cleveland1984graphical,heer2010crowdsourcing}, whereas the text of the prompts was referenced from the experimental materials of \cite{heer2010crowdsourcing}. Our study included seven tasks from these studies plus two new variants. Each task consisted of 45 distinct trials.

\textbf{Stimuli and Tasks.} To create the stimuli (i.e., visualizations), we first generated ten numerical values using the formula from \citet{cleveland1984graphical}:
\begin{align}
    s_i = 10 \times 10 ^{(i-1)/12}, i = 1, 2, ..., 10
\end{align}
We then constructed all 45 possible unique pairs of these values.
The ratios of these pairs ranged from $0.18$ to $0.83$.
For each of the seven tasks from the original studies \cite{cleveland1984graphical,heer2010crowdsourcing}, we generated 45 visualizations corresponding to these pairs.
In each visualization, the segments encoding the values being compared were colored blue and yellow and also labeled ``A'' and ``B'' (Figure \ref{fig:tasks}).
All other values in the visualization (i.e., values not being compared) were generated randomly, with a few constraints.
For instance, the bottom of the bar segments being compared in Task 4 had to be unaligned; otherwise, the perceptual task would essentially become identical to Task 2.

\section{Experiments 1 and 2}

\textbf {Experiment 1.} For each trial, the VLM was given a visualization and asked to respond to the probes:
\begin{enumerate}[leftmargin=*]
    \setlength\itemsep{0em}
    \setlength\parskip{0em}
    \setlength\parsep{-2em}
    \item Which of the two, blue (A) or yellow (B), shapes is smaller?
    \item What percentage is the SMALLER marked shape of the LARGER? Enter a \% between 0 and 100.
\end{enumerate}

We vary the framing of the prompts in two ways. The first is the explicit mention of color in the probe (no color/has color). The "has color" prompt contains references to the colors of the two labeled segments. The second is requesting explanations in the model response (no explanation/has explanation). The "has explanation" prompt asks the model not just to provide an answer (such as identifying which visual element is smaller or the proportion between two segments), but also to generate the reasoning behind its decision.
See Appendix \ref{sec:prompts} for the exact inputs to the VLM.

\begin{figure}[h!]
    \centering
    \subfigure[]{
        \includegraphics[width=0.55\textwidth]{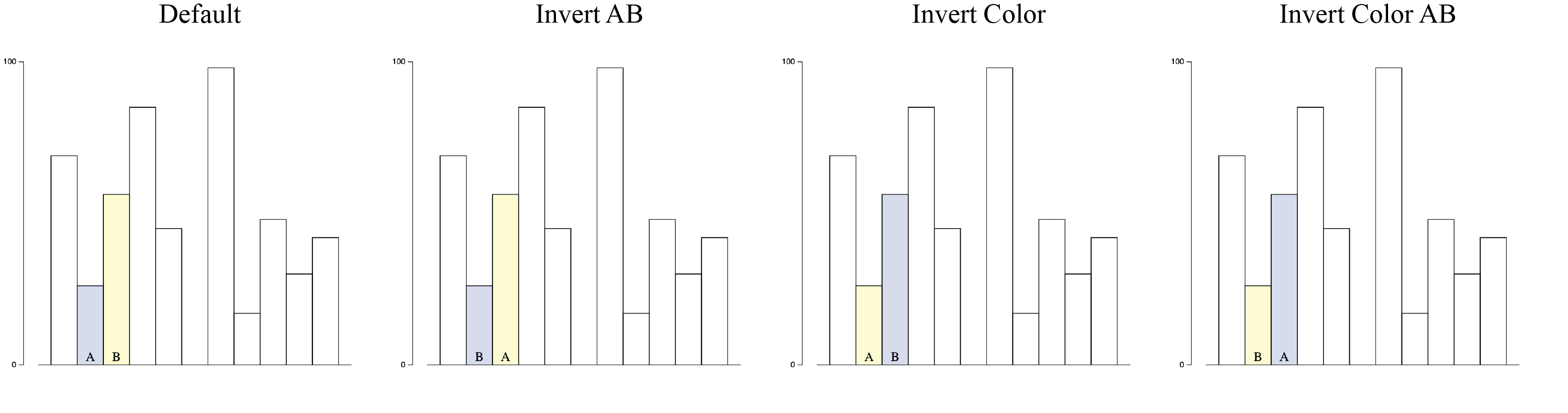}
    }\hfill
    \subfigure[]{
        \includegraphics[width=0.42\textwidth]{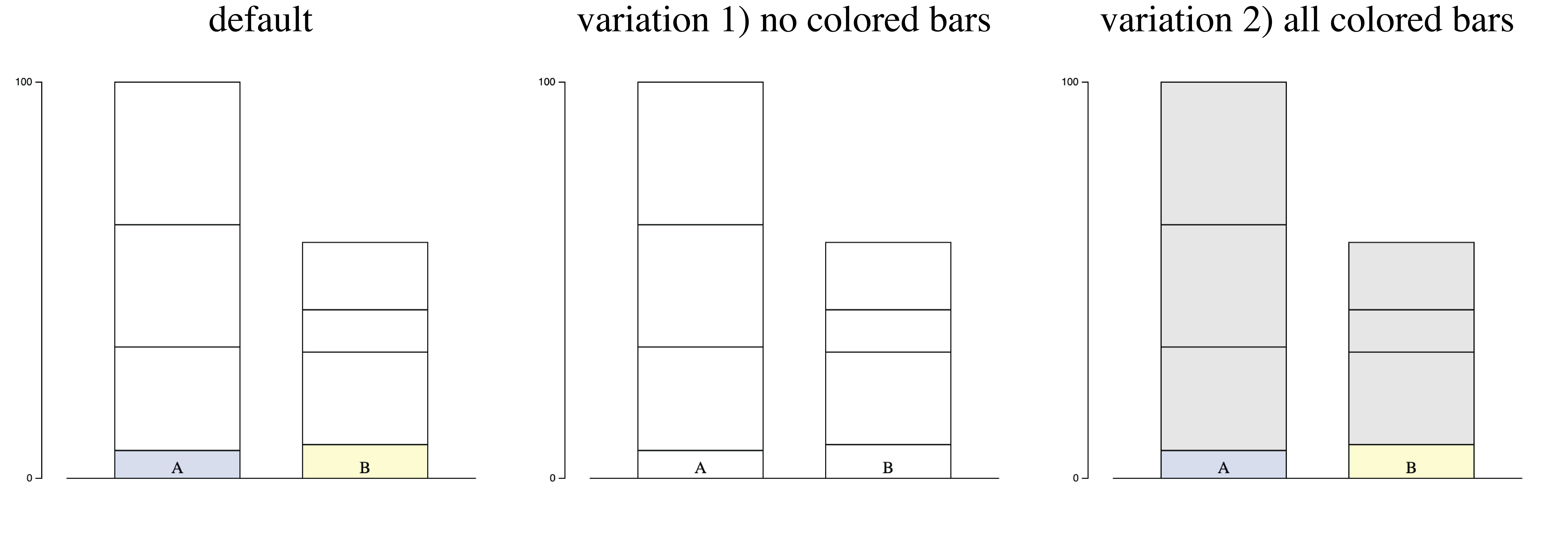}
    }
    \caption{(a) Experiment 1 inverted colors and AB labels.
    (b) Stimulus variations in Experiment 2.
    }
    \label{fig:study12}
\end{figure}

Finally, since VLMs may exhibit bias towards left/right layouts and A/B labels, we added three stimuli variations that inverted the order of colors and A/B labels (Figure \ref{fig:study12}a).

\textbf {Experiment 2.} In addition to the stimuli used in prior studies (i.e., Figure \ref{fig:tasks}), we created two new variations of all ($7 \times 45 =$) $315$ visualizations and evaluated VLM comprehension of these stimuli using the same probes and prompt framings. These variations are, first, no colored bars, and second, all colored bars; Figure \ref{fig:study12}b). Other than fill color changes, the stimuli in the variations were identical.

\subsection{Results}

\begin{table*}[!ht]
    \centering

    \resizebox{\textwidth}{!}{%
    \begin{tabular}{cc|c|cc|ccccccc|c|cc}
    \hline
          \multicolumn{2}{c|}{ Prompts} &  Stimuli & \multicolumn{2}{c|}{ Invert} & ~ & ~ & ~  & Tasks & ~ & ~ & ~ & ~ &\multicolumn{2}{|c}{ Data fit} \\ 
         \hline
         
        Clr. & Exp. & & AB & Clr. & Task 1 & Task 2 & Task 3  & Task 4 & Task 5 & Task 6 & Task 7 & Overall & $\rho$ & $p$ \\ \hline
        ~ & ~ & Def. & & & 1.00 & 0.89 &  1.00 &  0.74 & 0.69 & 0.75 & 0.63 & 0.81 & 0.90 & 0.006\\
        \xmark & \xmark & All clr. & \xmark & \xmark & 0.88 & 0.44 & 0.92 & 0.58 & 0.53 & 0.62 &  0.62 & 0.66 & 0.18 & 0.699 \\ 
        ~ & & No clr. & & &  0.53 &  0.44 & 0.44 &  0.53 & 0.53 & 0.38 & 0.44 & 0.47 & 0.19 & 0.679 \\ 
         \hline
         ~ & & Def. & & & 1.00 & 0.96 & 1.00 & 0.82 & 0.69 & 0.82 & 0.73 & 0.86 & 0.89 & 0.007 \\
        \xmark & \cmark & All clr. & \xmark & \xmark & 0.96 & 0.56 & 0.91 & 0.76 & 0.58 & 0.80 & 0.69 & 0.75 & 0.39 & 0.383\\ 
        ~ & & No clr. & & & 0.44 & 0.49 & 0.51 & 0.47 & 0.58 & 0.60 & 0.49 & 0.51 & 0.52 & 0.229\\ \hline
         ~ & & Def. & & & 1.00 & 0.82 & 1.00 & 0.76 & 0.82 & 0.71 & 0.71 & 0.83 & 0.73 & 0.060\\ 

       \cmark & \xmark & All clr. & \xmark & \xmark & 0.78 & 0.47 & 0.80 & 0.58 & 0.62 & 0.67 & 0.69 & 0.66 & 0.07 & 0.879\\ 
        
        ~ & & No clr. & & & 0.62 & 0.44 & 0.47 & 0.44 & 0.56 & 0.38 & 0.44 & 0.48  & 0.37 & 0.413\\ \hline
 
        ~ & & Def. & & & 1.00 & 0.91 & 1.00 & 0.75 & 0.73 & 0.91 & 0.73 & 0.87 & 0.84 & 0.017\\ 

        \cmark & \cmark & All clr. & \xmark & \xmark & 0.87 & 0.56 & 0.91 & 0.64 & 0.56 & 0.87 & 0.80 & 0.75 & 0.22 & 0.638\\ 
        
        ~ & & No clr. & & &  0.56 & 0.51 & 0.44 & 0.49 & 0.56 & 0.51 & 0.40 & 0.50 & 0.36 & 0.423\\ \hline \hline

        ~ & ~ & & \cmark & \xmark & 0.98 & 0.78	& 0.98 & 0.78 &	0.58 & 0.87 & 0.98 & 0.85 & 0.19 & 0.688  \\
        
        \cmark & \cmark & Def. & \xmark & \cmark & 1.00 & 0.93 & 1.00 & 0.82 & 0.73 & 0.58 & 0.69 & 0.82 & 0.83 & 0.021  \\
        
        ~ & ~ & & \cmark & \cmark & 1.00 & 0.83 & 1.00 & 0.74 & 0.71 & 0.86 & 1.00 & 0.88 & 0.22 & 0.632  \\
        
        \hline
        

    \end{tabular}
    }
    \caption{Accuracy of GPT-4o-mini on probe 1 ("smaller than") judgments. For each combination of prompt and stimuli variation, we calculate Spearman's rank correlation relative to human judgments reported in Figure 4 of \citet{heer2010crowdsourcing}, taking negative of $\rho$ due to the opposite rankings that log error and accuracy yield. \textbf{Prompt variations -} \textbf{Clr:} Has color; \textbf{Exp:} Has explanation; \textbf{Stimuli variations -} \textbf{Def:} Default stimuli; \textbf{All clr:} All color stimuli; \textbf{No clr:} No color stimuli. See Figure \ref{fig:study_variations}a for examples of \textbf{Def, All clr, and No clr}.
    \textbf{Inversion variations -} \textbf{AB:} Invert labels for A and B ; \textbf{Clr} Invert colors associated with A and B. }
    \label{tab:accuracies}
\end{table*}

We test both probes by Direct Probing to elicit VLM judgments, where we ask the VLM about its current state.

Table \ref{tab:accuracies} shows GPT-4o-mini accuracies on probe 1 across the seven tasks, four prompt framings, three stimuli variations, and three combinations of inverted color and A/B labels explored in experiments 1 and 2.
Note that for the inverted color and A/B labels conditions, we only looked at prompts with both color and requested explanations (\textbf{Has Color, Has explanation}), and the default stimuli condition (Figure \ref{fig:study12}b, left).

The key takeaways are as follows:

\textbf{Experiment 1 - Prompt Sensitivity:} Overall, the model performed best in the \textbf{Has Color, Has Explanation} prompt condition. Removing either cue from the prompt (either no color or no explanation) led to a small drop in model accuracy. Removing both color and explanations led to a substantial decline, as seen in the \textbf{No Color, No Explanation} condition.
This demonstrates that explicit mentions of color and requesting explanations play large roles in enhancing the model's graph comprehension.

\textbf{Experiment 1 - Color and Label Inversion:} Inverting color and A/B labels do not affect model performance, with overall model accuracies remaining high.
However, there is a decline in data fit when A/B labels are inverted.
We discuss this further in Section \ref{discussion}

\textbf{Experiment 2 - Input Stimuli:}
The model generally performed better on ``Default'' stimuli than ``All Color'' stimuli and better on these than ``No Color'' stimuli.
This suggests that model performance can be impacted through stylistic changes, even when the data and data mappings used are the same.

\section{Experiment 3}

\begin{figure}[H]
    \centering
    \begin{minipage}{0.45\textwidth}
        \centering
        \includegraphics[width=\linewidth]{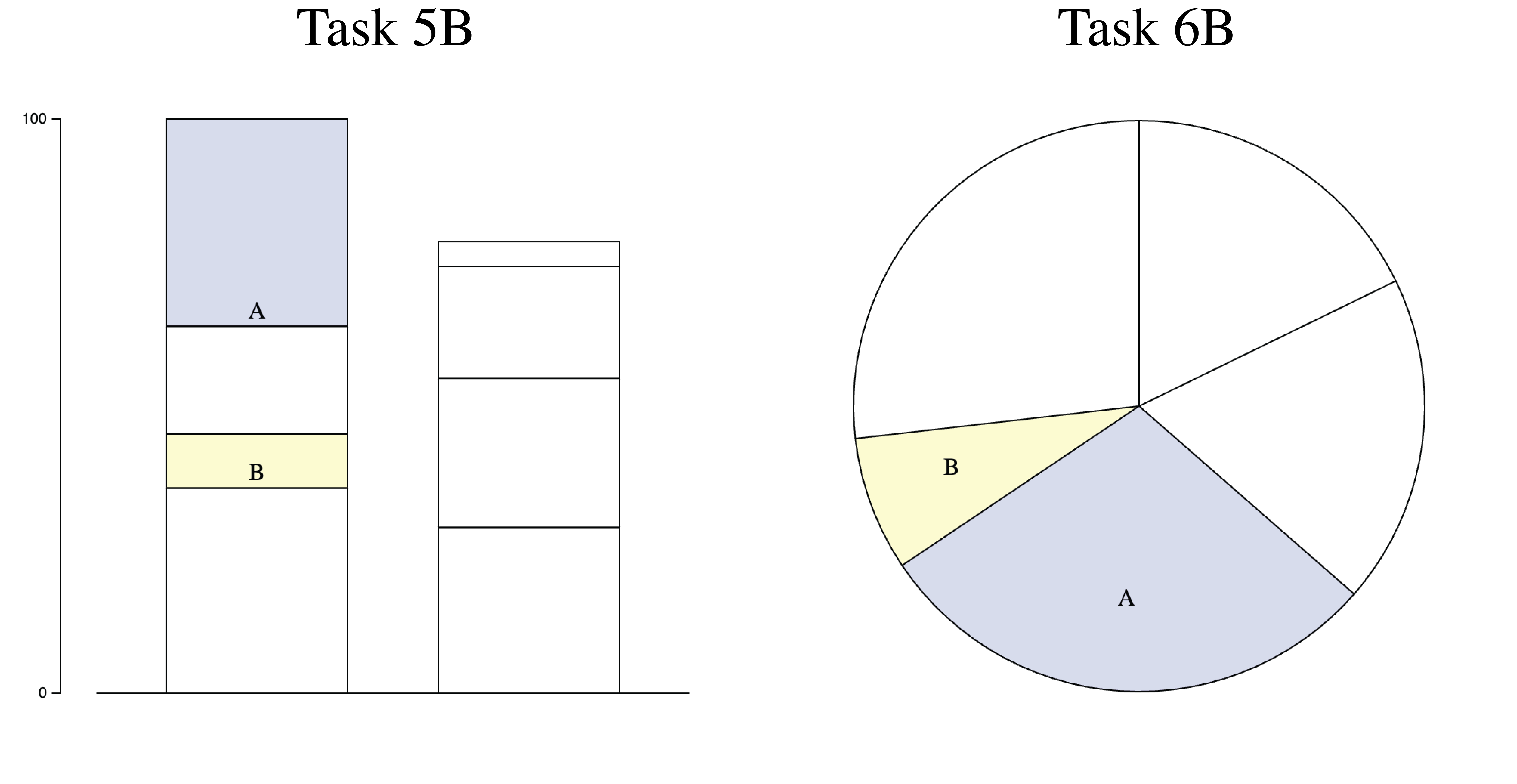}
        \caption{Task variations in Experiment 3.}
        \label{fig:study_variations}
    \end{minipage}\hfill
    \begin{minipage}{0.5\textwidth}
        \centering
        \resizebox{\textwidth}{!}{%
            \begin{tabular}{cc|c|cc|cccc}
                \hline
                \multicolumn{2}{c|}{Prompts} & Stimuli & \multicolumn{2}{c|}{Invert} & ~ & \multicolumn{2}{c}{Tasks} & ~ \\ 
                \hline
                Clr. & Exp. & & AB & Clr. & Task 5 & Task 5B & Task 6 & Task 6B \\ \hline
                \cmark & \cmark & Def. & \xmark & \xmark & 0.73 & 0.84 & 0.92 & 0.76\\ \hline \hline
                ~ & ~ & ~ & \cmark & \xmark & 0.58 & 0.64 & 0.87 & 0.83\\
                \cmark & \cmark & Def. & \xmark & \cmark & 0.73 & 0.71 & 0.58 & 0.82 \\ 
                ~ & & & \cmark & \cmark & 0.71 & 0.71 & 0.86 & 0.93\\ \hline
            \end{tabular}
        }
        \captionof{table}{Accuracy of GPT-4o-mini on probe 1 ("smaller than") judgments on Experiment 3 task variants. Task 5 and 6 accuracies were copied from Table \ref{tab:accuracies} for comparison.}        \label{tab:study3_accuracies}
    \end{minipage}
\end{figure}

\textbf {Experiment 3.} To disentangle the effect of contiguous segments on model performance, we created variations of Tasks 5 and 6 (henceforth 5B and 6B) that change whether the segments used for comparison are contiguous with one another (Figure \ref{fig:study_variations}b). In Task 5, the segments being compared are always contiguous, whereas in Task 5B, they are always separated by another segment. In Task 6, the segments are always separated by other segments, whereas in Task 6B, they are always contiguous.

Since Experiment 1 demonstrates that model accuracy is highest when the prompt framing includes color and explanation, we use this framing here for Experiment 3 as well.
Similarly, based on Experiment 2 results, we apply the best-performing default variant in this experiment (Figure \ref{fig:study_variations}b).

\subsection{Results}

    
         
        


\textbf{Experiment 3  - Segment Contiguity:} There was an effect of segment contiguity on model performance. For the default condition, the model was less accurate when the segments being compared were contiguous than when they were well-separated (VLM performance Task 5B $\gt$ Task 5; Task 6B~$\lt$~Task 6).
However, inverting segment colors also inverts this relationship, causing contiguous segments to perform better than separate segments.

\section{Discussion} \label{discussion}

\textbf{Comparison to human performance:} To evaluate the relationship between VLM performance and human performance, we conducted a rank-order correlation analysis (Table \ref{tab:accuracies}, Data fit). We ordered the difficulty of the seven tasks for the VLM by their accuracy on probe 1 and for humans by their log error values from \citet{heer2010crowdsourcing}.
(Note that these two approaches rank the results in descending and ascending order, respectively, so we take the negative value of calculated $\rho$.)
There is greatest correspondence between VLM and humans ($\rho = 0.90$) on the relative difficulty of the seven tasks for the default prompt framing (\textbf{No Color, No Explanation}) and the default stimulus presentation.
More broadly, there is a strong correlation across prompt variations for default stimuli.
Conversely, there is a consistently low correlation for "All Color" stimuli.

Interestingly, Experiment 1 suggests that there is an effect of label order on model correlation to human performance even when average model accuracies remain high.
Inverting the layout of A/B labels leads to a decline in data fit.
However, we do not see similar effects when color is inverted.




\begin{figure}[h!]
    \centering
    \subfigure[]{
        \includegraphics[trim={0cm 0cm 0cm 0cm}, width=0.48\textwidth]{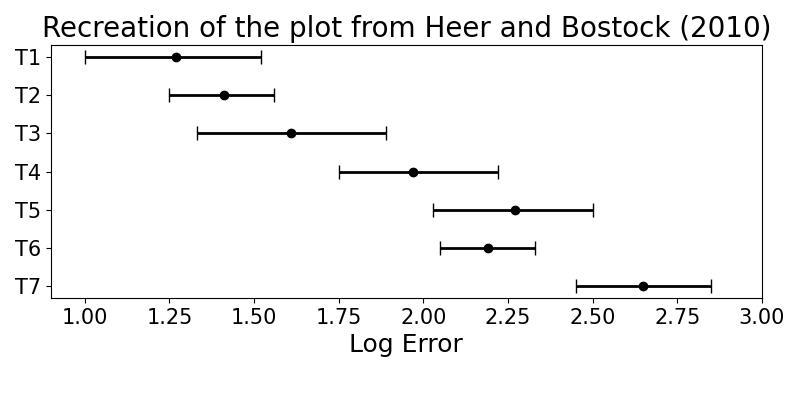}      
    }
    \subfigure[]{
        \includegraphics[trim={0cm 0cm 0cm 0cm}, width=0.48\textwidth]{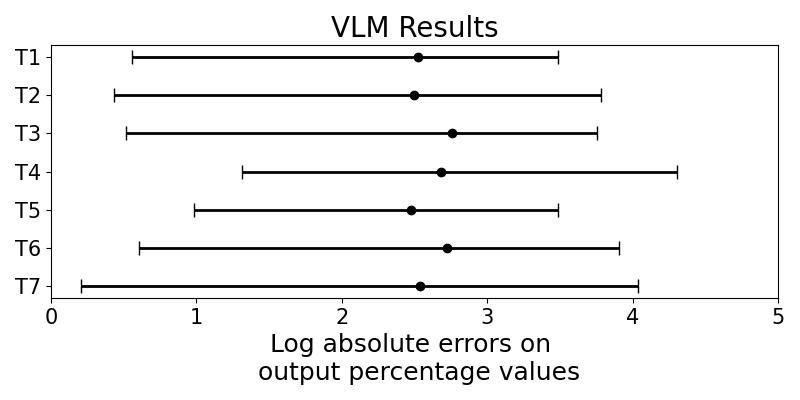}
    }
   
    \caption{Accuracy of VLMs on proportion judgments (probe 2).}
    \label{fig:logmae}
\end{figure}

In addition to probe 1 accuracies reported above, we also evaluate the accuracy of the proportion judgments (probe 2) by replicating \citet{heer2010crowdsourcing} and calculating the log absolute error $(\log_2(|\text{Judged} \% - \text{True} \%| + \frac{1}{8})$ and 95\%
confidence intervals for the seven tasks.
Note that unlike for probe 1, the VLM does not align with human performance. Whereas humans showed systematic differences in the accuracy of proportion judgments across the tasks (Figure \ref{fig:logmae}a), there was no statistical difference in the accuracy of the VLM (Figure \ref{fig:logmae}b).

\section{Conclusion}



This paper reports the initial findings of evaluating GPT-4o-mini in a zero-shot manner on graphical perception tasks with established human performance profiles \cite{cleveland1984graphical, heer2010crowdsourcing}. The study assesses the model's ability to extract and compare data from segments in a visualization.
Our results show that VLMs perform similarly to humans when 1) both color and explanations are present in the prompt template, 2) segments are colored in the visualization, and 3) segments are non-contiguous.
This suggests that, for certain combinations of task and visualization type, VLMs have the potential to design and evaluate visualizations by modeling human performance.

Looking ahead, the findings here may be useful for predicting and explaining VLM performance on more complex chart types, as seen in real-world applications.
For instance, the effect of segment contiguity, documented here in the novel comparisons between Task 5 and 5B and Task 6 and 6B, may result in lower accuracies on ChartQA tasks \cite{masry2022chartqa} for stacked bar charts and pie charts overall.
Future work can also evaluate human performance on the Task 5B and 6B variations introduced here to establish whether VLMs can generate new predictions about human performance on novel chart comprehension tasks.

\section{Acknowledgements}
The authors would like to thank the anonymous reviewers for their thoughtful and detailed feedback on the paper.
This work is supported in part by NIH grant 1U01CA284207.

\bibliographystyle{plainnat}
\bibliography{references}
\section{Appendix}

\section*{Input prompts to VLM}
\label{sec:prompts}
Here are the different kinds of prompts used in the study.

\subsection*{Prompt - No Color, No Explanation}
\begin{lstlisting}
Which of the two, (A) or (B), shapes is smaller?

 - When the inputs are bar charts or wedges compare the lengths/ height.
 - When the inputs are pie charts or circles compare the area.

Select one of the following:

A. The marked shape (A) is smaller.
B. The marked shape (B) is smaller.

What percentage is the SMALLER marked shape of the LARGER? Enter a percentage between 0 and 100.

Output in JSON format:
{
  "Is A smaller than B": true/false,
  "percentage": "XX%"
}
\end{lstlisting}

\subsection*{Prompt - Has Color, No Explanation}
\begin{lstlisting}
Which of the two, blue (A) or yellow (B), shapes is smaller?

 - When the inputs are bar charts or wedges compare the lengths/ height.
 - When the inputs are pie charts or circles compare the area.

Select one of the following:

A. The marked blue shape (A) is smaller.
B. The marked yellow shape (B) is smaller.

What percentage is the SMALLER marked shape of the LARGER? Enter a percentage between 0 and 100.

Output in JSON format:
{
  "Is A smaller than B": true/false,
  "percentage": "XX%"
}
\end{lstlisting}

\subsection*{Prompt - No Color, Has Explanation}
\begin{lstlisting}
Which of the two, (A) or (B), shapes is smaller?

 - When the inputs are bar charts or wedges compare the lengths/ height.
 - When the inputs are pie charts or circles compare the area.

Select one of the following:

A. The marked shape (A) is smaller.
B. The marked shape (B) is smaller.

What percentage is the SMALLER marked shape of the LARGER? Enter a percentage between 0 and 100.

Output in JSON format:
{
  "explanation for smaller or bigger": "...",
  "Is A smaller than B": true/false,
  "explanation for percentage": "...",
  "percentage": "XX%"
}
\end{lstlisting}

\subsection*{Prompt - Has Color, Has Explanation}
\begin{lstlisting}
Which of the two, blue (A) or yellow (B), shapes is smaller?

 - When the inputs are bar charts or wedges compare the lengths/ height.
 - When the inputs are pie charts or circles compare the area.

Select one of the following:

A. The marked blue shape (A) is smaller.
B. The marked yellow shape (B) is smaller.

What percentage is the SMALLER marked shape of the LARGER? Enter a percentage between 0 and 100.

Output in JSON format:
{
  "explanation for smaller or bigger": "...",
  "Is A smaller than B": true/false,
  "explanation for percentage": "...",
  "percentage": "XX%"
}
\end{lstlisting}

\section*{Limitations and Future Work}

We acknowledge a few limitations of our work. Our analysis used only one VLM, limiting the findings' generalizability, as different VLMs may exhibit varying performance characteristics. 
In particular, we expect that VLMs fine-tuned for chart comprehension or chart question-answering tasks will outperform general-purpose models like GPT-4o-mini.
Future work should thus consider testing multiple VLMs to create a more comprehensive evaluation. 
We also observed significant uncertainty in the model's performance on tasks involving percentage judgments, which indicates lower model performance on these proportion-type judgments. Further testing would be useful to better understand and potentially mitigate this uncertainty.

Another limitation is that the input stimuli used for these experiments may not resemble the types of visualizations found in the training data of GPT-4o-mini. This mismatch could have contributed to suboptimal performance in specific tasks. Future studies can modify the input stimuli to match visualization styles in the training data better to evaluate model accuracy and reliability more precisely.

\appendix

\end{document}